# A Multi-tiered Human-in-the-loop Approach for Interactive School Mapping Using Earth Observation and Machine Learning

*Casper Fibaek [a], Abi I. Riley [b, c], Kelsey Doerksen [b, d], Do-Hyung Kim [b], Rochelle Schneider [a]*

[a] European Space Agency, ESRIN, Via Galileo Galilei, 1, 00044 Frascati Roma, Italy

[b] UNICEF-ITU, GIGA Initiative, Place des Nations, 1211 Geneva 20, Switzerland

[c] Imperial College London, 90 Wood Ln, London W12 0BZ, UK

[d] University of Oxford, Wolfson Building, Parks Rd, Oxford OX1 3QD, UK

## Abstract

This paper presents a multi-tiered human-in-the-loop framework for interactive school mapping designed to improve the accuracy and completeness of educational facility records, particularly in developing regions where such data may be scarce and infrequently updated. The first tier involves a machine learning based analysis of population density, land cover, and existing infrastructure compared with known school locations. The first tier identifies potential gaps and "mislabelled" schools. In subsequent tiers, medium-resolution satellite imagery (Sentinel-2) is investigated to pinpoint regions with a high likelihood of school presence, followed by the application of very high-resolution (VHR) imagery and deep learning models to generate detailed candidate locations for schools within these prioritised areas. The medium-resolution approach was later removed due to insignificant improvements. The medium and VHR resolution models build upon global pre-trained steps to improve generalisation. A key component of the proposed approach is an interactive interface to allow human operators to iteratively review, validate, and refine the mapping results. Preliminary evaluations indicate that the multi-tiered strategy provides a scalable and cost-effective solution for educational infrastructure mapping to support planning and resource allocation.

# 1. Introduction

Ensuring every school is accurately mapped is key to bridging the digital divide (Doerksen et al., 2024), improving educational outcomes worldwide, and achieving the Sustainable Development Goals (Anderson et al., 2017). Accurate and up-to-date school maps support governments and non-governmental organisations in allocating resources effectively, planning infrastructure and internet connectivity, and ultimately empowering communities through improved access to educational opportunities (Vijil-Morin et al., 2023). Improved connectivity has been shown to increase educational outcomes (UNESCO, 2023).

Existing geospatial datasets of schools often rely heavily on data collections that may be incomplete, inaccurate, or outdated. School datasets frequently have missing schools, contain imprecise coordinate locations, and feature inconsistent metadata (Riley et al., 2025). A common occurrence is to have the number of schools at the administrative level but without further coordinates. Poor datasets result in an incomplete picture of the education infrastructure landscape, particularly in areas with high poverty, conflict levels, or rapidly developing regions (Govinda, 1999; Tingzon et al., 2020). Traditional mapping methods, like manual image interpretation or field surveys, can be time-consuming and costly. Another approach is using very high-resolution (VHR) satellite imagery with automated detection to offer a solution (Maduako et al., 2022). However, it may become prohibitively expensive for large-scale, global projects, especially if frequent updates are necessary.

Earth Observation (EO) provides a valuable source of timely, free, and global data that can be used for consistent and comparable information to support the school mapping process. Recent advancements in machine learning (ML), especially deep learning, have significantly improved the potential for automating mapping from EO data (Vargas-Munoz et al., 2020).

A central part of the framework is the integration of a human-in-the-loop process. Automated workflows sometimes miss contextual clues or generate false positives, especially in heterogeneous settings. A WebGIS-based interactive interface enables human experts, such as local education officials, to continually review, validate, and refine the results produced by the deep learning models, thereby improving trust in the produced datasets.

Our research introduces a multi-tiered human-in-the-loop framework to address the challenges of cost, scale, and accuracy in school mapping. By strategically integrating multiple data sources, including population maps, land cover information, building footprints, medium-resolution Sentinel-2 imagery, and nightlight data, with targeted use of VHR imagery, our approach reduces dependency on expensive VHR datasets. The framework has broad applications beyond school mapping, and the source code is provided using open-source licenses.

## 1.1 Related Work

Mapping educational infrastructure, particularly schools, is essential for resource allocation and achieving development goals, like ensuring inclusive and equitable quality education (SDG 4 - Quality Education). Early efforts often relied on Geographic Information Systems (GIS) and participatory methods (PPGIS) to compile school location inventories (Govinda, 1999; Vijil-Morin et al., 2023). However, these approaches frequently struggle with data completeness and timeliness, especially in dynamic environments or regions with limited resources.

The increasing availability of EO data and advancements in machine learning, particularly deep learning, have introduced automated methods for mapping the built environment (e.g., Qiu et al., 2020). Researchers have applied convolutional neural networks (CNNs) and other architectures like Vision Transformers (ViTs) to extract buildings from satellite and aerial imagery, sometimes fusing multiple data sources like LiDAR and optical imagery or integrating EO data with existing vector data like OpenStreetMap (Huang and Zhang, 2019; Vargas-Munoz et al., 2020).

Recent studies demonstrate the potential of deep learning applied to very high-resolution (VHR) satellite imagery to map schools specifically. Maduako et al. (2022) used a CNN to identify schools across diverse geographical contexts, finding that models trained on varied regional data generalise better than country-specific models. Doerksen et al. (2024) employed weakly supervised ensemble classifiers (Vision Transformers and CNNs) on image tiles labelled only for school presence, achieving high precision and using Class Activation Mapping (CAM) (Selvaraju et al., 2017) for locating schools within tiles. One issue with this approach is that our preliminary findings show that the CAM maps often highlight school indicators, such as playgrounds, as schools instead of school buildings.

Recognising the limitations of single data sources, many mapping frameworks now integrate multi-modal data. Medium-resolution data like Sentinel-2 has proven useful for delineating broader human settlement extents (Qiu et al., 2020). Data on nighttime light emission from VIIRS can be a valuable proxy for human activity, electrification, and development. Using these nightlight datasets makes it possible to detect small rural settlements (Elvidge et al., 2017). The density of the population correlates with the presence of infrastructure like schools. While VHR imagery is crucial for identifying individual structures (Sirko et al., 2021), the acquisition cost often makes it necessary to target applications. Fusing different data types, such as LiDAR and spectral imagery for building detection, can significantly improve accuracy by reducing false positives (Huang and Zhang, 2019), supporting the rationale for multi-tiered data fusion approaches. However, LIDAR data is generally more costly than VHR imagery.

Integrating human expertise through human-in-the-loop (HITL) systems is gaining traction in EO analysis. Interactive methods where human labellers refine model outputs or systems that flag uncertain predictions for human review (García Rodríguez et al., 2020) can improve accuracy efficiently and improve trust in mapping systems. This finding aligns with principles from participatory mapping, where local knowledge enhances geospatial data (Govinda, 1999). Modern tools like MapSwipe enable large-scale volunteer contributions for validating EO-derived maps (Ullah et al., 2023).

Embedding techniques, generating vector representations from data, were explored for school mapping and connectivity prediction (Doerksen et al., 2024). While Vision Transformers (ViTs) were benchmarked against CNNs for mapping, fine-tuned CNNs often performed better and ran faster, which is important for interactive mapping. For connectivity prediction, geographically aware location encoders (Fibaek et al., 2024) were tested, but engineered features derived from multi-modal data generally yielded higher accuracy. Combining embeddings with engineered features showed some potential, suggesting complementarity but highlighting the current advantage of domain-specific features for this task (Doerksen et al., 2024).

This work builds on the shift towards integrating multi-modal EO data, advanced machine learning, and human oversight for infrastructure mapping. Here, challenges remain in developing scalable and cost-effective systems specifically for comprehensive school inventories, particularly in resource-constrained settings. Our research addresses this gap by proposing and evaluating a multi-tiered, interactive framework that combines multiple EO data sources (population density, land cover, Sentinel-2, nightlights, VHR) with machine learning and a human-in-the-loop validation and fine-tuning interface. This approach aims to use the strengths of each data source and methodology tier while taking steps to mitigate their limitations. The goal is to directly respond to the need for improved educational infrastructure data outlined in the introduction.

## 2. Study Area and Data Sources

### 2.1 Study Area Description

To evaluate the robustness and adaptability of our framework across diverse geographical, infrastructural, and socioeconomic contexts, this study encompasses the entire continent of Africa. Africa presents a wide spectrum of environments, from densely populated urban centres to vast, sparsely inhabited rural areas, and varying levels of existing infrastructure documentation. This diversity and need for educational infrastructure mapping make Africa a good location for testing a scalable mapping methodology designed to handle heterogeneous conditions and data availability.

The quality and completeness of existing school location data vary significantly across the continent. While some regions, particularly parts of Eastern Africa, may have more established or higher-quality datasets, many areas suffer from under-documented educational infrastructure, outdated records, or data gaps, especially in remote or conflict-affected zones. Applying the framework continent-wide allows us to assess the performance across varied conditions and its potential to contribute to an improved understanding of educational facility distribution across Africa.

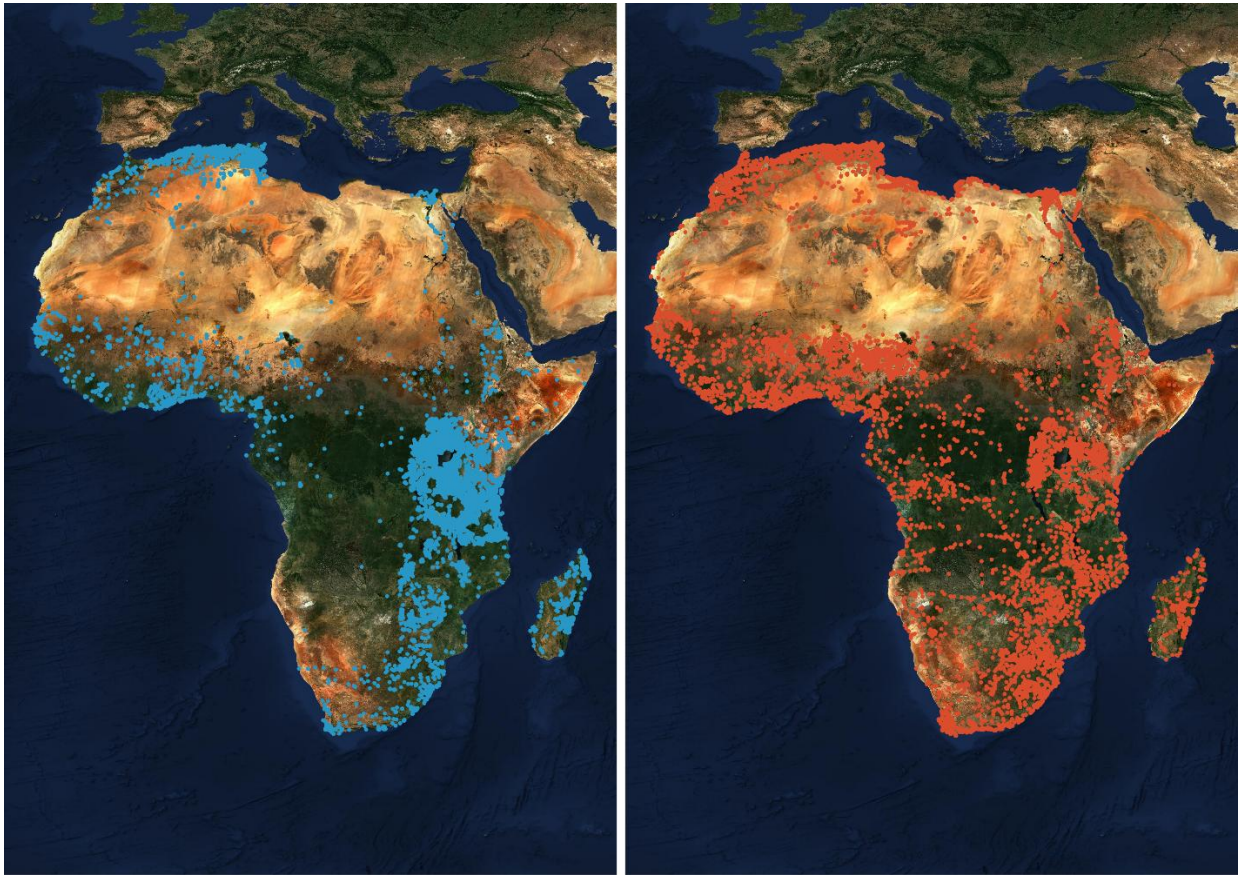

*Figure 1. Study area of Africa with school (blue) and non-school locations (red). Basemap: Sentinel 2 Cloudless by EOX.*

## 2.2 Data Sources

The multi-tiered approach relies on integrating geospatial datasets from multiple modalities. By combining information from different sources, we can lower the impact of the weaknesses of the individual datasets. Broad-scale datasets like GHSL population data and VIIRS nightlights provide the initial context, indicating areas of human settlement where schools are likely to exist. Building footprint datasets (Microsoft, Google, OSM) offer crucial information on the presence and location of structures, helping to filter potential school locations and provide negative samples. Medium-resolution Sentinel-2 imagery was investigated to analyse settlement patterns and land cover over wider areas, while targeted VHR imagery enabled the detailed visual inspection necessary for identifying specific school buildings. Known school locations (UNICEF, OSM) and non-school POIs (OSM) serve as essential ground truth for training and validating the models. At the same time, static datasets (WorldCover, Terrain, Climate) provide a stable environmental context.

### *UNICEF School Dataset*

A core dataset of the presented work is the UNICEF School Dataset (Tatarinov & Ambos, 2025). This database contains information on school locations, names, and metadata. It is the primary source for known school locations for training, validation, and identifying potential gaps. Data quality varies by country, necessitating preprocessing and validation steps detailed in the methodology chapter.

### *OpenStreetMap (OSM)*

OpenStreetMap (Vargas-Munoz et al., 2020) provides crowd-sourced, open-access geographic vector data. It provides valuable contextual data, including building footprints, road networks, and points of interest (POIs). It can supplement official school datasets or provide negative samples (non-school buildings) for model training. We retrieved school and non-school POIs (e.g., hospitals, churches, markets) from OSM (Herfort et al., 2023).

### *Global Human Settlement Layer (GHSL)*

The Global Human Settlement Layer (GHSL) project produces global datasets on human presence and the built environment derived from satellite imagery and census data (European Commission, 2023). We use several GHSL products (R2023A release):

- GHS-POP: Population density grids.
- GHS-BUILT-S: Built-up surface fraction.
- GHS-SMOD (DEGURBA): Settlement model classifying areas into cities, towns, suburbs, and rural areas based on population density and contiguity (European Commission et al., 2021). These layers provide crucial context for Tier 1 analysis.

### *Microsoft Building Footprints*

The Microsoft Building Footprints dataset (Microsoft, 2023) provides building footprints extracted from satellite imagery using deep learning. This dataset filters out erroneous location points potentially falling in uninhabited areas and provides building context for VHR analysis.

*Google Open Buildings Dataset*

The Google Open Buildings dataset (Sirko et al., 2021) provides building footprints across Africa extracted from high-resolution satellite imagery. The dataset is an additional source for validating school locations or identifying potential building candidates in areas with limited OSM coverage. The Microsoft and Google Building datasets were merged with the OSM building dataset to form a single building dataset. The approach prioritised OSM buildings, with high-confidence buildings from Microsoft and Google added in places where OSM had no buildings available.

*Sentinel-2*

Sentinel-2, a constellation of two satellites from the European Space Agency (ESA), provides high-resolution (10-60 m) multispectral optical imagery with frequent revisit times (approx. 5 days). Its 10 m spatial resolution bands can be used to identify medium-scale settlement patterns and land cover changes. It was investigated for Tier 2 analysis to narrow down search areas identified in Tier 1. Sentinel 2 and 1 are used to calculate the GHSL data (European Commission, 2023).

*VIIRS Nightlights*

The Visible Infrared Imaging Radiometer Suite (VIIRS) nightlight data from the Suomi National Polar-orbiting Partnership ([Suomi NPP](#)) measures nighttime light emissions, offering insights into human activity, electrification, and economic development (Elvidge et al., 2017). With a spatial resolution of roughly 500m (processed), it serves as a valuable proxy for identifying inhabited areas, particularly in the first tier of the analysis, complementing population and settlement data.

*Very High Resolution (VHR) Imagery*

VHR imagery (sub-meter resolution) from commercial providers like Maxar is crucial for the detailed analysis in Tier 3. This level of detail allows for identifying and classifying individual buildings or small structures that might be schools. Due to cost, VHR imagery is used selectively within our framework, targeted towards high-probability areas identified by earlier tiers. For our experiments, we used predominantly Maxar RGB imagery.

*Static Reference Datasets*

Several static global datasets were used to provide environmental context. The ESA WorldCover dataset classified global land cover at 10m resolution (Zanaga et al., 2022). Global terrain characteristics were derived from the Iwahashi and Yamazaki (2022) dataset, which classifies terrain based on slope and basin characteristics. Finally, climate context was incorporated using the Köppen–Geiger climate classification system, which categorises global climate zones (Kottek et al., 2017). These datasets provide stable background information against which dynamic features like population and settlement patterns can be analysed.

## 3. Methodology

The proposed framework employs a multi-tiered approach to progressively refine the search space for identifying potential school locations, culminating in human validation. The approach can be conceptualised as a 'funnel' and is described in Table 1 below.

| Tier | Objective | Primary Data | Core Method |
|---|---|---|---|
| 1 | Identify areas with unexpected school distributions | Population, Settlement (GHSL), Buildings (Microsoft/Google), OSM, Nightlights | Random Forest Model |
| 2 | Narrow search space to high-impact regions *(Later removed)* | Sentinel-2 | Convolutional Neural Network |
| 3 | Generate specific school candidates | VHR Imagery | Convolutional Neural Network |
| 4 | Refine and validate candidate locations | VHR Imagery, Candidate List | Interactive Human-in-the-loop Interface |
| 5 | In-person validation | Data from previous tiers. | Field visit |

*Table 1. Overview of Processing Tiers*

Tier two was eventually removed from the process, as it did not significantly improve the final results while increasing processing time and complexity.

### 3.1 Data Preprocessing and Sample Creation

Prior to beginning the mapping process, the input data needed to be cleaned and normalised. In addition to data cleaning, a new dataset of negative samples needed to be created to allow the training of binary classifiers.

*3.1.1 School Dataset Preparation and Validation*

Cleaning the school data dataset involved multiple steps:

> (1). **Deduplication**: Merging points of schools likely to refer to the same school. To accomplish this step, schools closer than 25m to each other with a similar name were merged. The merging was done using fuzzy string matching with a Levenshtein similarity of at least 85%, following a preprocessing of case-folding and accent removal (Riley et al., 2025). In some cases, this can merge primary and secondary schools at the same location; however, for the sake of this study, these were considered the same school.

(2). **Missing Coordinates**: In some cases, no coordinates were available for a given school; however, an address and name might still be available. Using geocoding, an attempt was made to rectify this and add coordinates to schools (Riley et al., 2025). If the returned geocoded address was not located in the same administrative zone as specified for the school, the school was removed from the dataset.

(3). **Geographic filtering**: For each of the school points, the distance to the nearest building (in the combined OSM (Vargas-Munoz et al., 2020), Google (Sirko et al., 2021), and Microsoft (Microsoft, 2023) dataset) was calculated along with the land cover classification derived from the ESA World Cover dataset (Zanaga et al., 2022). Schools located in water bodies or more than 150m from a registered building were removed from the dataset.

(4). **Stratification:** TTo reduce the size of the dataset and align it with the number of negative samples, 10000 school points were chosen based on a stratified sampling of the DEGURBA (European Commission et al., 2021) classes and a minimum of 10 km distance to other schools in the dataset.

This process aimed to create the most accurate and comprehensive baseline school location dataset possible for the study area.

### 3.1.2 Creation of Positive and Negative Samples for Schools

For training binary classifiers of schools, it is possible to rely on learning from positive and unlabeled data (PU learning), where the unlabelled data mixes positive and negative samples (Bekker & Davis, 2020). However, for this study, we set out to generate a complementary set of known negative schools to train the binary classifiers. Generating known negatives of schools is difficult as buildings in many places could be schools despite not having labels. To generate the known negatives, the following approach was taken:

(1). **Points of Interest**: The initial step was to find points of interest that could be schools but explicitly are not schools. To achieve this, we used the OSM overpass API (Olbricht, 2024) to find POIs that were explicitly not schools. Some of the features we identified were shops, tourism, offices, military, leisure, healthcare, and churches. A complete list is available in the accompanying repository. All POIs where school or school-related terms were used in the name were excluded for negative sampling. To create the list of excluded words, a list of school synonyms and names for schools in the local language was produced and used to exclude PoIs.

(2). **Metadata filtering**: Following inspection of the identified points in the step above, it became clear that, in general, places with a name attached to them were of higher quality. That meant they had more metadata and were more likely to be located within buildings from other datasets. Due to this, all points that did not explicitly have a name were excluded from the dataset.

(3). **Geographical Filtering**: Similar to the step in the cleaning of the positive samples, points that were located within water bodies in the ESA WorldCover dataset (Zanaga et al., 2022) were not included along with points not located within the joined (OSM (Vargas-Munoz et al., 2020), Google (Sirko et al., 2021), Microsoft (Microsoft, 2023)) buildings layer.

(4). **Stratification**: 8000 Points of Interest were chosen randomly and stratified based on the DEGURBA classification (European Commission et al., 2021).

(5). **Non-urban negatives**: To allow the models to learn that non-inhabited areas do not contain schools, 2000 negative samples were generated for places further than 1km away from a structure based on the JRC build-up mapping, which is based on sentinel 1 and 2 (Fibæk et al., 2022). For stratification, these 2000 points were randomly chosen using the ESA WorldCover dataset (Zanaga et al., 2022).

These points were used as the basis for extracting VHR and Sentinel 2 imagery. The dataset is available in the repository.

## 3.2 Tier 1 – Machine Learning Based Expectation Modelling

The main objective of Tier 1 is to identify areas where the observed distribution of schools significantly deviates from an expected pattern derived from population density, settlement characteristics, and environmental factors. Finding these discrepancies allows for identifying regions with incomplete school records (fewer schools than expected) or containing anomalies requiring further investigation. The tier uses a Random Forest (RF) model (scikit-learn developers, 2024), building upon the approach detailed by Riley et al. (2025).

The RF models were trained using a diverse set of geospatial features extracted for known school locations (positive samples) and non-school locations (negative samples; see Section 3.1.2). The input features included:

- **Geographic Coordinates:** Encoded using sine and cosine transformations to capture spatial patterns.
- **Climate Zone:** Derived from the Köppen–Geiger climate classification (Kottek et al., 2006).
- **Land Cover:** Based on the ESA WorldCover 10m dataset (Zanaga et al., 2022).
- **Terrain Type:** Classified using the global terrain polygons dataset (Iwahashi and Yamazaki, 2022).
- **Population Density:** Primarily using GHSL population grids (European Commission, 2023), potentially supplemented by methods similar to Fibæk et al. (2022).
- **Settlement Type:** Using the GHSL Settlement Model Degree of Urbanisation (DEGURBA) classification (European Commission et al., 2021; European Commission, 2023).

- **Nighttime Lights:** VIIRS Night Lights data as a proxy for human activity and electrification (Elvidge et al., 2017).

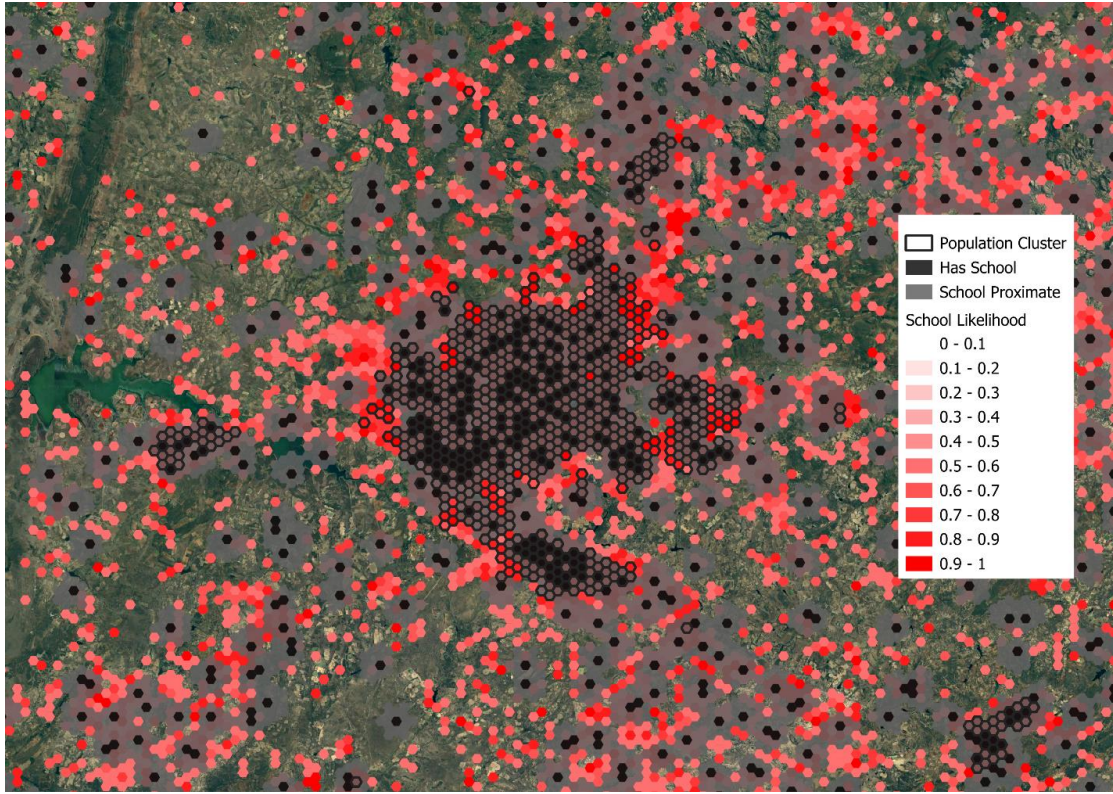

*Figure 2. Example of tier 1 results showing predicted school likelihood overlaid with known population clusters and existing school locations. High-likelihood areas with few known schools and a high population indicate potential gaps for investigation.*

The models were implemented using the scikit-learn library's Random Forest classifier. Hyperparameter tuning was performed using a randomised search strategy on a 20% subset of the training data. The classification model's performance and the input features' relative importance are summarised in Tables 2 and 3.

| Class | Precision | Recall | F1-Score |
| --- | --- | --- | --- |
| School (1) | 0.92 | 0.88 | 0.90 |
| Non-School (0) | 0.88 | 0.92 | 0.90 |

*Table 2: Classification performance metrics for the Tier 1 Random Forest model.*

| Feature | Importance |
| --- | --- |
| Coordinates | 0.233 |
| Climate | 0.019 |
| Degurba | 0.225 |
| Landcover | 0.038 |

| Feature | Importance |
|---|---|
| Terrain | 0.031 |
| Population | 0.358 |
| Nightlights | 0.096 |

*Table 3: Feature importance scores for the Tier 1 Random Forest model.*

The model achieved a balanced F1-score of 0.90, indicating strong performance in distinguishing between school and non-school locations based on the input features. As expected, features directly related to human presence and settlement density (Population, Degurba, Nightlights) accounted for most (approximately 2/3) of the feature importance. Geographic coordinates also played a significant role (23%), likely due to the model learning spatial patterns associated with urbanisation and infrastructure development not fully captured by the other variables. Climate, land cover, and terrain features had lower, but still non-negligible, importance.

The output of this tier is a map indicating the likelihood of school presence across the study area (as exemplified in Figure 2). Areas with high predicted probability but few known schools are prioritised for investigation in the subsequent tiers.

### 3.3 Tier 2 – Narrowing Search Space with Medium-Resolution Imagery

The initial design of the framework included a second tier aimed at further refining the priority areas identified in Tier 1 using medium-resolution Sentinel-2 imagery. The objective was to leverage the broader spatial context and spectral information available in Sentinel-2 data (10m resolution) to pinpoint specific 1km$^2$ zones within the Tier 1 priority regions that showed characteristics strongly associated with school presence before resorting to generally costly VHR analysis.

To achieve this, a deep learning approach was explored. A convolutional neural network (CNN) was intended to classify 256x256 pixel (10m) Sentinel-2 image tiles based on their likelihood of containing a school. A pre-training step was undertaken using a large-scale, geographically diverse dataset to enhance the model's generalisability across diverse global environments. We utilised the MajorTOM Sentinel-2 core dataset (Francis and Czerkawski, 2024), a comprehensive collection of Sentinel-2 imagery covering the Earth's land surface.

Given the large size of the MajorTOM dataset, a subset termed "FastTOM" was created. This subset consisted of smaller 256x256 pixel tiles (derived from the original 1068x1068 patches) and excluded tiles with over 50% cloud or water cover, resulting in a ~3TB dataset suitable for pre-training. The pre-training employed a geography-aware self-supervised learning strategy (Ayush et al., 2021), where the model learns representations by predicting the geographic location of the input image tile.

Following pre-training, the model was fine-tuned using the prepared positive (school) and negative (non-school) samples described in Section 3.1.2, extracting corresponding Sentinel-2 imagery for each location.

However, experimentation revealed that this Sentinel-2 based classification tier did not significantly improve predictive performance over the baseline probabilities generated by the Tier 1 Machine Learning model. The features discernible at Sentinel-2's 10m resolution proved insufficient to reliably differentiate between areas likely containing schools and other similar settlement patterns within the already prioritised regions. The tests showed that some marginal gains were possible in areas but did not justify the substantial increase in computational cost and workflow complexity associated with processing and analysing Sentinel-2 data at this scale. Consequently, Tier 2 was removed from the final operational framework, and the high-probability areas identified directly from Tier 1 were used to guide the VHR analysis in Tier 3. This highlights the challenge of distinguishing specific building functions like schools using only medium-resolution satellite imagery, even with advanced deep learning techniques.

A topic of future research could be adding automatically extracted image embeddings for the sentinel 2 patches using global foundation models, to the Random Forest model of tier one.

## 3.4 Tier 3 – VHR Candidate Generation with Deep Learning

Within the newly defined priority areas from Tier 2, VHR imagery was analysed using deep learning models to identify specific building candidates likely to be schools. This tier builds on the detailed VHR classification experiments from Doerksen et al. (2024).

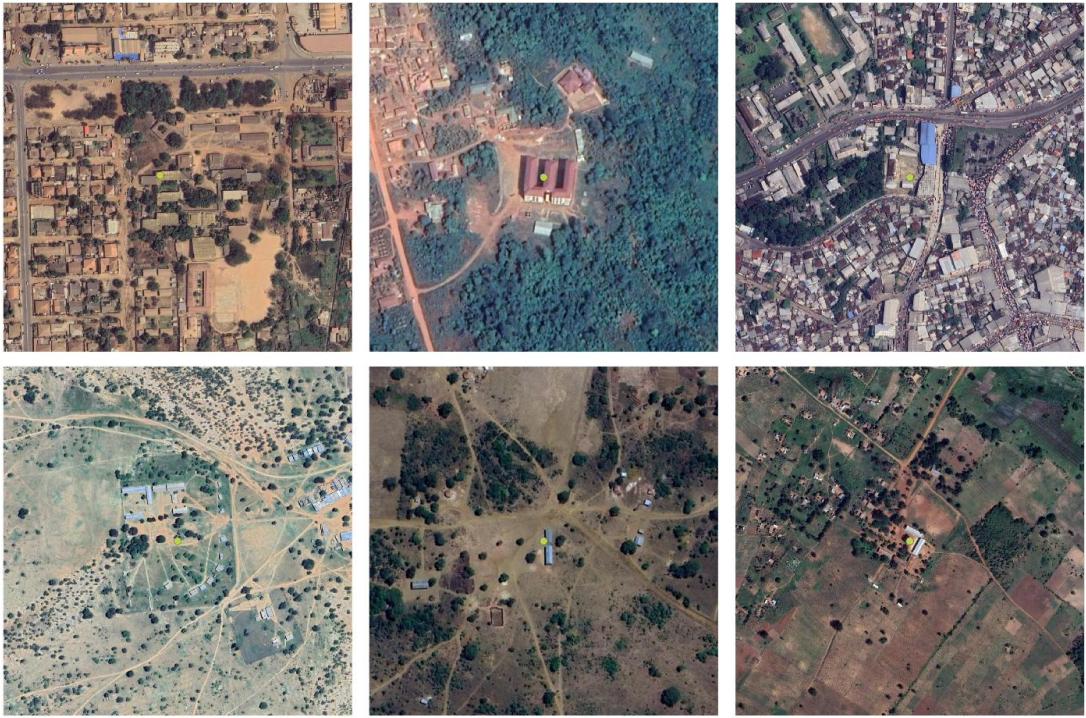
*Figure 4: Examples of Schools within the dataset*

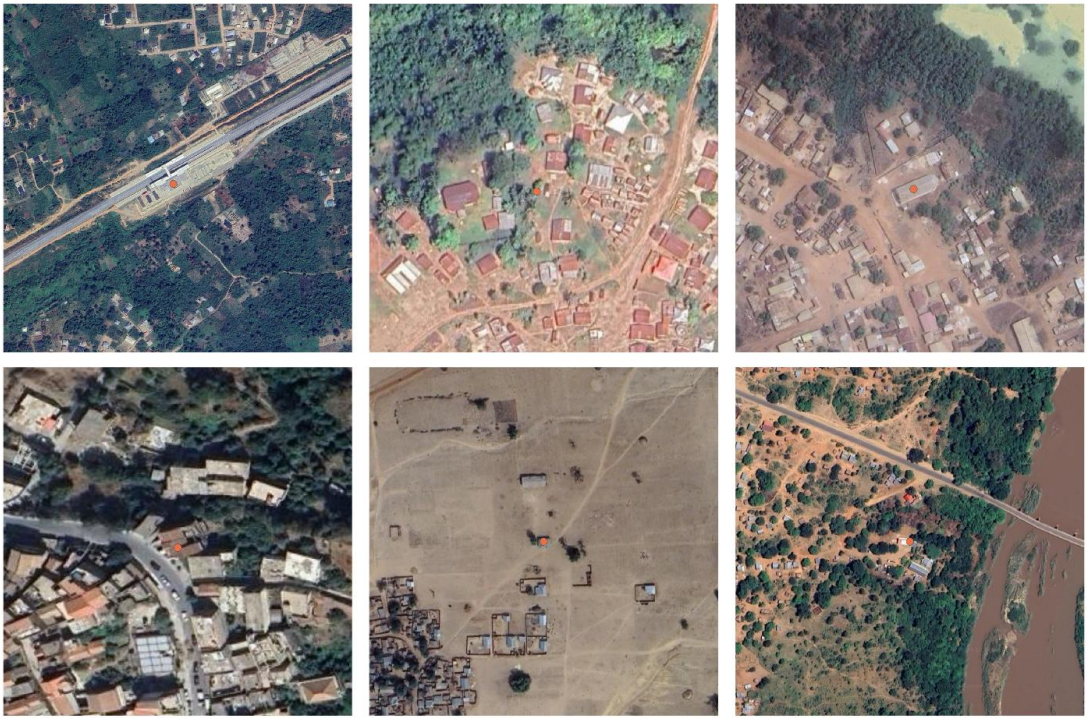
*Figure 5: Examples of non-schools within the dataset.*

The VHR classification models build on two models with the same general architecture but trained in separate ways:

### 3.4.1 Foundation Model

The foundation model is trained across a large and diverse dataset of VHR imagery that spans the globe. The locations were chosen based on stratified sampling, with at least one sample of every city in the world with above 100,000 inhabitants. Following the initial sampling of cities, a stratified sampling of climate zones was performed. Inhabited places were chosen based on the GHSL settlement model (European Commission et al., 2021) and sampled twice as frequently as uninhabited places.

The model itself is based on the ConvNext (Liu et al., 2022) model, a convolutional neural network that has been shown to perform well on VHR imagery. The model was trained using the same approach as Doerksen et al. (2024), except that the model was trained on 256x256 pixel tiles instead of 512x512 pixel tiles. The model was trained using semi-self-supervised learning, which was trained on predicting the encoded coordinates and the class of global static layers (land cover, terrain, and climate). The latent space was constrained using a cosine similarity loss based on augmentations to ensure the model learns a meaningful representation of the data. The model was trained using a batch size of 128 and a learning rate of 0.0001.

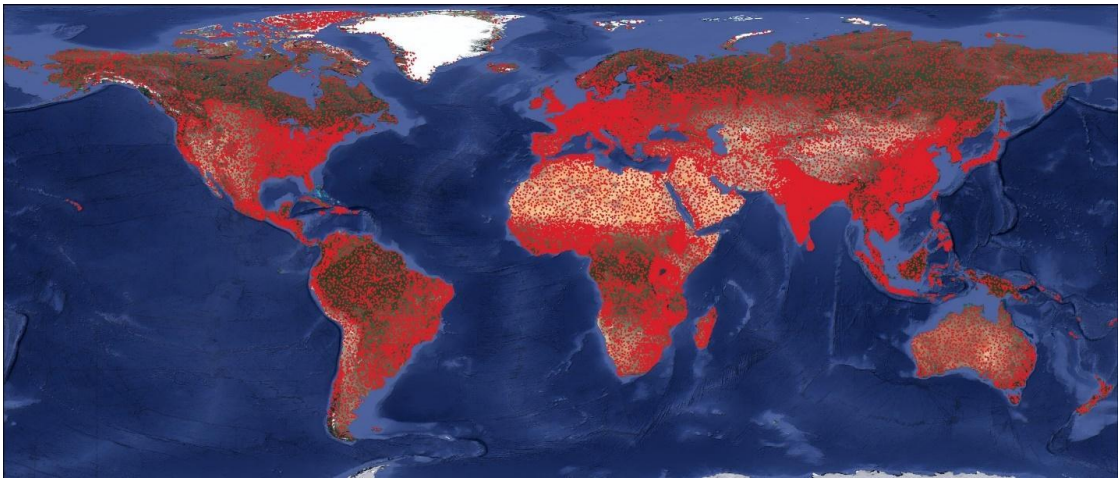

*Figure 6. Global sampling of VHR imagery*

### 3.4.2 Local Model

The local model was based on the foundation model, using its weight as the initial starting point. It was trained using simple binary cross-entropy loss on the positive and negative samples described in Section 3.1.2. The model was trained using a batch size of 128 and a learning rate of 0.0001. Image augmentations were created using the Buteo library (Fibaek, 2025).

The local model, fine-tuned on the positive and negative samples specific to the study area, achieved a classification accuracy of **83.2%**. This figure represents the model's performance in distinguishing between VHR image tiles containing schools and those containing non-school structures or empty areas, based on the features learned during both the foundation pre-training and the subsequent local fine-tuning phase. While this accuracy indicates a strong capability for automated identification, it also underscores the need for the subsequent human-in-the-loop validation tier to address the remaining ~17% of misclassifications (false positives and false negatives). It is important to note that the model is applied to areas where the Tier 1 model has indicated a high probability of school presence, as the model was trained on a dataset that is biased towards populated areas.

### 3.5 Tier 4 – Human-in-the-Loop Validation and Refinement

The final tier involved human experts reviewing the candidate list generated by Tier 3 using an interactive interface. This step is important for validating automated detections, correcting errors, and incorporating local knowledge.

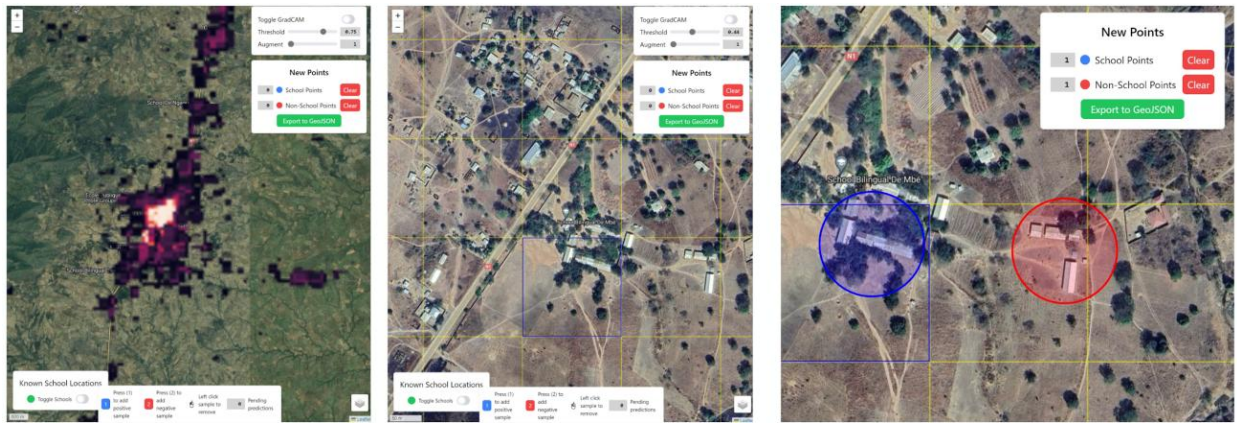

*Figure 7: Screenshots of the Interface*

The interface is based on a WebGIS interface, built on top of a Flask (Ronacher et al., 2024) backend in Python. The interface is designed for interactive exploration and validation of deep learning model predictions on satellite imagery. The interface dynamically fetches map tiles from standard Web Map Service (WMS) providers corresponding to the user's viewport. The map is created using LeafletJS (Agafonkin et al., 2023). For each requested tile, the backend preprocesses the images into tensors suitable for the deep learning 'local' PyTorch model (Paszke et al., 2017). Test-time augmentation techniques, including flips and rotations, are applied to enhance prediction robustness before feeding the data into the model.

The model outputs a probability score indicating the likelihood of the target feature (e.g., a school) being present within the tile. These predictions are calculated on the fly and returned to the frontend as JSON data, allowing for near real-time visualisation of predictions being overlaid on the base map. Performance is optimised through multiple caching layers: HTTP caching for fetched WMS tiles and in-memory caching for recent prediction results.

A key interface feature is the integration of Gradient-weighted Class Activation Mapping (Grad-CAM), specifically GradCAM++ (Selvaraju et al., 2017). Users can activate this feature for any given tile via a request parameter. When enabled, the backend calculates the Grad-CAM heatmap for the specific tile using predefined target layers within the CNN architecture. This heatmap, highlighting the image regions most influential in the model's prediction, is then rendered as a semi-transparent PNG image and sent back to the client, overlaying the satellite imagery. This allows the operator to inspect parts of the model's reasoning process visually. Furthermore, the interface loads and displays ground truth data points (e.g., known school locations from GeoJSON files) alongside the model's predictions, enabling the operator to make direct comparisons and validate them. The interface is designed to support a human-in-the-loop workflow where user feedback on prediction accuracy (marking predictions as true or false) can be collected, forming a basis for subsequent model fine-tuning and improvement iterations.

This human validation step draws inspiration from interactive systems described by Buscombe et al. (2022), Rodríguez et al. (2020), and the validation tools used by Sivasubramanian et al. (2024) and Tingzon et al. (2020).

## 4. Results

This section summarises key findings related to the performance of the different tiers of the mapping framework, drawing from the methodology section. Tier two is left out, as the initial results proved unable to improve upon the tier one results.

### 4.1 Tier 1: Machine Learning Modelling Performance

The Machine Learning Random Forest model, detailed in Section 3.2, demonstrated strong performance in distinguishing potential school locations from non-school locations based on broad-scale geospatial features. As shown in Table 2, the model achieved a balanced F1-score of 0.90, indicating the model's ability to identify areas where the observed school distribution deviates from expectations derived from factors like population density, settlement type (Degurba), nighttime lights, and geographic coordinates, which were the most important features (Table 3). This tier successfully prioritised regions likely containing unmapped schools or anomalies, fulfilling its objective of guiding subsequent, more detailed analysis, building upon the approach validated by Riley et al. (2025). The output map (Figure 2) effectively highlights areas with high predicted probability but few known schools, serving as input for Tier 3.

### 4.2 Tier 3: VHR School Classification Performance

The fine-tuned local ConvNext-based model, described in Section 3.4.2, was applied to VHR imagery within the priority areas identified by Tier 1. This model, leveraging pre-training on a global foundation dataset and fine-tuning on specific positive and negative samples, achieved a classification accuracy of 83.2% in distinguishing VHR image tiles containing schools from those without. This performance, consistent with findings from related work using similar deep learning approaches on VHR data for school mapping (Doerksen et al., 2024), confirms the feasibility of using deep learning on VHR imagery to generate a high-confidence list of school candidates suitable for validation in the next tier.

The model's accuracy is reasonably high, but residual errors emphasise the usefulness of the subsequent human review.

### 4.3 Tier 4: Human-in-the-Loop Component Evaluation

While quantitative results for the Tier 4 interface are part of ongoing work, preliminary assessments and related literature support its value. The manual review conducted during the initial dataset preparation (Section 3.1.1) demonstrated the need for human oversight to correct errors inherent in existing datasets and preliminary automated extractions. Furthermore, existing research consistently shows significant accuracy improvements when human validation is incorporated into Earth Observation mapping workflows. For instance, studies like those by García Rodríguez et al. (2020) and Tingzon et al. (2020) report substantial gains in metrics like the F1 score through human-in-the-loop refinement. The focused nature of the validation, where human experts review only the high-probability or uncertain candidates generated by Tier 3 within priority areas identified by Tier 1, enhances the efficiency of expert time usage. Interfaces similar to the one developed (Figure 7), even simple ones, allow for rapid review, as demonstrated in crowdsourcing applications like MapSwipe (Ullah et al., 2023). Therefore, the human-in-the-loop stage is expected to significantly improve the precision and overall reliability of the final school map by effectively filtering false positives generated by automated methods and potentially identifying missed schools by applying expert knowledge and local context.

## 5. Discussion

This study introduced and evaluated a multi-tiered, human-in-the-loop framework for interactive school mapping. The results demonstrate the potential of the framework to aid in creating accurate and complete school inventories.

### 5.1 Effectiveness of the Multi-tiered Approach

The tiered structure effectively narrowed the search space for potential schools, functioning as a funnel from broad-scale analysis to fine-scale verification. Tier 1 successfully leveraged readily available, coarse-resolution datasets (population, settlement, nightlights) and EO-based ML modelling to identify large regions where school data was likely incomplete or different to expectations, efficiently focusing resources. An intermediate Tier 2, designed to use medium-resolution Sentinel-2 imagery for further refinement between Tier 1 and Tier 3, was explored but ultimately removed from the framework. As detailed in Section 3.3, experimentation showed that the 10m resolution of Sentinel-2 provided insufficient detail to reliably improve the localisation of potential schools beyond what Tier 1 achieved, and the marginal benefits in areas did not justify the added complexity and processing time. Tier 3 then applied computationally intensive deep learning models to costly VHR imagery only within these prioritised areas (potentially with a school), generating specific building candidates. This progressive refinement strategy effectively balances computational cost, data acquisition cost (particularly for VHR imagery), and spatial coverage, making large-scale mapping more feasible. The successful integration of data across different scales and types in Tiers 1 and 3 aligns with findings from other multimodal mapping studies, confirming that such integration generally improves mapping outcomes.

## 5.2 Contribution of the Human-in-the-loop Component

While the Tier 3 automated classification model achieved respectable accuracy (83.2%), the inclusion of Tier 4, the human-in-the-loop validation, remains critical to the framework's success. Automated methods, including deep learning models such as ConvNext, occasionally produce false positives and false negatives. This occurs due to factors such as variations in school architecture and surrounding environments, image quality issues, occlusions, or similarities between schools and other building types (Maduako et al., 2022).

The human-in-the-loop component addresses some of these limitations. While an expert might not be able to validate every identified school, an expert can rule out clearly non-schools and determine if it is worthwhile to continue on-the-ground validation. Firstly, it significantly improves precision as expert reviewers can effectively filter out false positives identified by the automated classifier. Secondly, it has the potential to improve recall, as local experts using the interactive interface (Figure 7) might identify schools missed by the algorithm by leveraging contextual knowledge not available to the model. Thirdly, involving local stakeholders, such as local education officials, in the validation process through the interactive tool builds trust and increases ownership of the resulting map. Finally, local human reviewers are likely better equipped to handle ambiguities, interpret complex scenes, and adapt to regional variations in school appearance than purely automated systems. However, these improvements in automated systems are moving fast, and this might no longer be true. This integration of automated detection with human expertise aligns with principles from participatory GIS (Govinda, 1999) and resonates with the successes reported in other human-in-the-loop EO applications (García Rodríguez et al., 2020; Tingzon et al., 2020).

## 5.3 Implications for Educational Planning and SDGs

The primary output of this framework, more accurate and comprehensive school location maps, has practical implications for educational development. It provides data to support initiatives like GIGA, a UNICEF and International Telecommunication Union (ITU) project, in planning and monitoring the efforts to connect every school to the internet by 2030 (Tatarinov & Ambos, 2025). Accurate maps enable the effective identification of underserved areas, allowing for better allocation of resources for building or improving infrastructure. Furthermore, precise school locations are important inputs for disaster management planning.

## 5.4 Limitations and Future Work

The performance of all tiers is inherently dependent on the quality, availability, and consistency of the input data, including official school records, VHR satellite imagery, and ancillary datasets such as building footprints, whose characteristics can vary significantly across different regions and countries. While the human validation in Tier 4 is important, its scalability remains a challenge when applied to very large areas or entire countries. Validating potentially thousands of candidates still requires a significant amount of human effort and resources. Future work could focus on improving the efficiency of the human-in-the-loop workflow, perhaps by using model uncertainty scores to prioritise the most ambiguous candidates for human review or by exploring carefully managed crowdsourcing approaches (akin to MapSwipe, Ullah et al., 2023) for initial filtering before expert validation.

The current version of the framework produces a static map representing school locations at a specific point in time. Incorporating time-series analysis, using archives of Sentinel-2 or Very High Resolution (VHR) imagery, would enable the monitoring of school construction, closures, or other changes over time. Additionally, the framework currently focuses solely on identifying the location of schools. Future extensions to the framework could extract additional attributes from the imagery, such as directly estimating connectivity, school size, population, or presence of recreational facilities. Validating the framework's output through on-the-ground validation campaigns with local partners is necessary to quantify real-world accuracy and refine the methodology accurately.

## 6. Conclusion

This research presented a multi-tiered, human-in-the-loop framework for interactive school mapping, specifically designed to address the global need for accurate and complete educational facility data. By strategically integrating Machine Learning modelling using broad-area datasets (Tier 1), deep learning classification applied to targeted Very High-Resolution Earth observation imagery (Tier 3), and an essential interactive human validation stage (Tier 4), our approach offers a scalable, cost-effective, and accurate solution compared to purely manual or purely automated methods alone. The machine learning models successfully identified priority areas with potential school data gaps based on population, settlement, and other geographic factors (F1-score 0.90). Within these areas, deep learning models demonstrated strong capability in classifying potential schools from VHR imagery (83.2% accuracy).

The human-in-the-loop component, implemented through an interactive web-based interface, is important for validating the automated detections and ensuring the reliability of the final maps. This framework provides a practical tool for governments, international organisations, NGOs, and initiatives like Giga to create and maintain the foundational school inventories necessary for effective planning of infrastructure development, including internet connectivity, equitable resource allocation, disaster response, and monitoring progress towards SDG 4. Accurate school maps are fundamental to improving educational opportunities for children worldwide. While future work can further refine aspects like validation efficiency and attribute extraction, this study demonstrates a significant advancement in applying integrated Earth observation and machine learning techniques for global educational development.

## Declaration of generative AI and AI-assisted technologies in the writing process

During the preparation of this work the author(s) used Grammarly for proofreading. After using this tool/service, the author(s) reviewed and edited the content as needed and take(s) full responsibility for the content of the publication.

Settlements using Machine Learning and Time Series Satellite Images: An Application in the Venezuelan Migration Crisis. https://doi.org/10.1109/AI4G50087.2020.9311041

Ullah, T., Lautenbach, S., Herfort, B., Reinmuth, M., Schorlemmer, D., 2023. Assessing Completeness of openstreetmap Building Footprints Using mapswipe. ISPRS Int. J. Geo-Inf. 12, 143. Https://doi.org/10.3390/ijgi12040143

UNESCO, 2023. Global Education Monitoring Report Summary 2023: Technology in education: A tool on whose terms? Paris, UNESCO. Available from: www.unesco.org/gemreport

Vargas-Munoz, J.E., Srivastava, S., Tuia, D. And Falc˜ao, A.X., 2020. Openstreetmap: Challenges and Opportunities in Machine Learning and Remote Sensing. *IEEE Geoscience and Remote Sensing Magazine*-PREPRINT. https://doi.org/10.1109/MGRS.2020.2994107

Vijil-Morin, A., Godwin, K., Ramirez, A., Mackintosh, A., mcburnie, C. And Haßler, B., 2023. *School mapping and decision-making*. https://doi.org/10.54676/EJZH8821

R. R. Selvaraju, M. Cogswell, A. Das, R. Vedantam, D. Parikh and D. Batra, "Grad-CAM: Visual Explanations from Deep Networks via Gradient-Based Localization," *2017 IEEE Int. Conf. Comput. Vis. (ICCV),* Venice, Italy, 2017, pp. 618-626, doi: https://doi.org/10.1109/ICCV.2017.74

Zanaga, D., Van De Kerchove, R., Daems, D., De Keersmaecker, W., Brockmann, C., Kirches, G., Wevers, J., Cartus, O., Santoro, M., Fritz, S., Lesiv, M., Herold, M., Tsendbazar, N.E., Xu, P., Ramoino, F., Arino, O., 2022. ESA worldcover 10 m 2021 v200. Https://doi.org/10.5281/zenodo.7254221

## Acknowledgements

This research was supported by the European Space Agency (ESA) Φ-lab and the United Nations Children's Fund (UNICEF) through the Giga initiative. We thank our partners for providing school location data and the contributors to OpenStreetMap, GHSL, Microsoft Building Footprints, and other open datasets used in this study.

## Code Availability

https://github.com/casperfibaek/school_mapping